\documentclass[11pt,a4paper]{article}
\usepackage[hyperref]{naaclhlt2018}
\usepackage{times}
\usepackage{latexsym}
\usepackage{graphicx}
\usepackage{mathtools}
\usepackage{tabularx}
\usepackage{amsmath}
\usepackage{booktabs} % For formal tables
\usepackage{url}
\usepackage{array}
\usepackage{ctable}
\usepackage{epstopdf}
\usepackage{hyperref}
\newcolumntype{P}[1]{>{\centeringarraybackslash}p{#1}}
\usepackage{url}

\newcommand\BibTeX{B{\sc ib}\TeX}

\title{Can Network Embedding of Distributional Thesaurus be Combined with Word Vectors for Better Representation?}

\author{Abhik Jana \\
  IIT Kharagpur, India \\
  {\tt abhik.jana@iitkgp.ac.in} \\\And
  Pawan Goyal \\
  IIT Kharagpur, India \\
  {\tt pawang@cse.iitkgp.ac.in} \\}

\date{}

\begin{document}
%\blindtext
\maketitle
\begin{abstract}
Distributed representations of words learned from text have proved to be successful in various natural language processing tasks in recent times. While some methods represent words as vectors computed from text using predictive model (Word2vec) or dense count based model (GloVe), others attempt to represent these in a distributional thesaurus network structure where the neighborhood of a word is a set of words having adequate context overlap.  Being motivated by recent surge of research in network embedding techniques (DeepWalk, LINE, node2vec etc.), we turn a distributional thesaurus network into dense word vectors and investigate the usefulness of distributional thesaurus embedding in improving overall word representation.     %Most of the attempts are made to represent words in intensional form directly from text whereas effect of an intensional representation prepared from an extensional representation is rarely investigated. In this paper, we investigate the outcome of converting a distributional thesaurus (extensional representation) into dense word vectors (intensional representation)  
%To be precise, we apply several network embedding methods like  etc. on distributional thesaurus network and evaluate the  word representation against word similarity and relatedness tasks.
This is the first attempt where we show that combining the proposed word representation obtained by distributional thesaurus embedding with the state-of-the-art word representations helps in improving the performance by a significant margin when evaluated against NLP tasks like word similarity and relatedness, synonym detection, analogy detection. Additionally, we show that even without using any handcrafted lexical resources we can come up with representations having comparable performance in the word similarity and relatedness tasks compared to the representations where a lexical resource has been used.
\end{abstract}
\section{Introduction}
Natural language understanding has always been a primary challenge in natural language processing (NLP) domain. \iffalse Lexical items and the interactions between them help learners interpret the text meaningfully. \fi Learning word representations is one of the basic and primary steps in understanding text and there are predominantly two views of learning word representations.
% ~\cite{chalmers2002sense}. %Intensional representation deals with meanings learned directly from reading text; basically any property or quality connoted by a word.  per intensional viewpoint 
In one realm of representation, words are vectors of distributions obtained from analyzing their contexts in the text and two words are considered meaningfully similar if the vectors of those words are close in the euclidean space. In recent times, attempts have been made for dense representation of words, be it using predictive model like Word2vec~\cite{mikolov2013efficient} or count-based model like GloVe~\cite{pennington2014glove} which are computationally efficient as well.
%l representation helps learners to learn how to choose words in a particular context. Specifically, 
Another stream of representation talks about network like structure where two words are considered neighbors if they both occur in the same context above a certain number of times. The words are finally represented using these neighbors. Distributional Thesaurus network is one such instance of this type, 
%extensional representation of words, 
the notion of which was used in early work about distributional semantics~\cite{grefenstette2012explorations,lin1998automatic,curran2002improvements}. \iffalse Earlier, preparing such sparse representations were computation intensive but With the advancement of technology researchers have tried to improve the efficiency of computation of sparse count-based methods~\cite{kilgarriff2004itri,biemann2013text}.\fi One such representation is JoBimText proposed by~\citet{biemann2013text} that contains, for each word, a list of words that are similar with respect to their bigram distribution, thus producing a network representation. Later,~\citet{riedl2013scaling} introduced a highly scalable approach for computing this network. We mention this representation as a DT network throughout this article. 
With the emergence of recent trend of embedding large networks into dense low-dimensional vector space efficiently~\cite{Perozzi:2014:DOL:2623330.2623732,tang2015line,grover2016node2vec} which are focused on capturing different properties of the network like neighborhood structure, community structure, etc., we explore representing DT network in a dense vector space and evaluate its useful application in various NLP tasks.
%of word similarity and relatedness. 
\iffalse We further explore whether the word representation learned throguh DT embedding can be meaningfully combined with the word representations of the first type (e.g., word2Vec, Glove), and whether the combination leads to performance gain, indicating the complimentary information being captured in both.
%In order to get dense vector representation of out of this DT network we apply network embedding methods~\cite{Perozzi:2014:DOL:2623330.2623732,tang2015line,grover2016node2vec} \\
Researchers have not been restricted to learning word embedding from raw text only; embeddings learned from images, other manually built lexical resources like WordNet~\cite{miller1995wordnet} have also been getting  appreciation in the community. Several attempts have been made to combine the representations by following multimodal distributional semantics~\cite{bruni2014multimodal} as well. Some researchers have used retrofitting~\cite{faruqui:2014:NIPS-DLRLW} as combination technique whereas others use vector concatenation, principal component analysis~\cite{jolliffe1986principal}, canonical correlation analysis~\cite{faruqui2014improving} etc. In this study, we restrict ourselves to the representations learned from raw text only.\fi

There has been attempt~\cite{ferret2017turning} to turn distributional thesauri into word vectors for synonym extraction and expansion but the full utilization of DT embedding has not yet been explored. In this paper, as a main contribution, we investigate the best way of turning a Distributional Thesaurus (DT) network into word embeddings by applying efficient network embedding methods and analyze how these embeddings generated from DT network can improve the representations generated from prediction-based model like Word2vec or dense count based semantic model like GloVe. %by evaluating extensively on different NLP tasks. 
We experiment with several combination techniques and find that DT network embedding can be combined with Word2vec and GloVe to outperform the performances when used independently. Further, we show that we can use DT network embedding as a proxy of WordNet embedding in order to improve the already existing state-of-the-art word representations as both of them achieve comparable performance as far as word similarity and word relatedness tasks are concerned. Considering the fact that the vocabulary size of WordNet is small and preparing WordNet like lexical resources needs huge human engagement, it would be useful to have a representation which can be generated automatically from corpus. We also attempt to combine both WordNet and DT embeddings to improve the existing word representations and find that DT embedding still has some extra information to bring in leading to better performance when compared to combination of only WordNet embedding and state-of-the-art word embeddings. While most of our experiments are focused on word similarity and relatedness tasks, we show the usefulness of DT embeddings on synonym detection and analogy detection as well. In both the tasks, combined representation of GloVe and DT embeddings shows promising performance gain over state-of-the-art embeddings.  
%For synonym detection the combined representation of GloVe and DT embeddings proved to be outperforming GloVe whereas for analogy detection DT embeddings itself shows promising performance beating GloVe.
%\input{Related_Work.tex}
\section{Related Work}
The core idea behind the construction of distributional thesauri is the distributional hypothesis~\cite{firth1957synopsis}: ``You should know a word by the company it keeps''. The semantic neighbors of a target word are words whose contexts overlap with the context of a target word above a certain threshold. Some of the initial attempts for preparing distributional thesaurus are made by~\citet{lin1998automatic},~\citet{curran2002improvements},~\citet{grefenstette2012explorations}. The semantic relation between a target word and its neighbors can be of different types, e.g., synonymy, hypernymy, hyponymy or other relations~\cite{adam2013evaluer,budanitsky2006evaluating} which prove to be very useful in different natural language tasks. Even though computation of sparse count based models used to be inefficient, in this era of high speed processors and storage, attempts are being made to streamline the computation with ease. One such effort is made by~\citet{kilgarriff2004itri} where they propose Sketch Engine, a corpus tool which takes as input a corpus of any language and corresponding grammar patterns, and generates word sketches for the words of that language and a thesaurus. Recently, ~\citet{riedl2013scaling} introduce a new highly scalable approach for computing quality distributional thesauri by incorporating pruning techniques and using a distributed computation framework. They prepare distributional thesaurus from Google book corpus in a network structure and make it publicly available.

In another stream of literature, word embeddings represent words as dense unit vectors of real numbers, where vectors that are close together in euclidean space are considered to be semantically related. In this genre of representation, one of the captivating attempt is made by~\citet{mikolov2013efficient}, where they propose Word2vec, basically a set of two predictive models for neural embedding whereas ~\citet{pennington2014glove} propose GloVe, which utilizes a dense count based model to come up with word embeddings that approximate this. Comparisons have also been made between count-based and prediction-based distributional models~\cite{baroni2014don} upon various tasks like relatedness, analogy, concept categorization etc., where researchers show that prediction-based word embeddings outperform sparse count-based methods used for computing distributional semantic models. In other study,~\citet{levy2014neural} show that dense count-based methods, using PPMI weighted co-occurrences and SVD, approximates neural word embeddings. Later,~\citet{levy2015improving} show the impact of various parameters and the best performing parameters for these methods. All these approaches are completely text based; no external knowledge source has been used.

More recently, a new direction of investigation has been opened up where researchers are trying to combine knowledge extracted from knowledge bases, images with distributed word representations prepared from text with the expectation of getting better representation. Some use Knowledge bases like WordNet~\cite{miller1995wordnet}, FreeBase~\cite{bollacker2008freebase}, PPDB~\cite{ganitkevitch2013ppdb}, ConceptNet~\cite{speer2017conceptnet}, whereas others use ImageNet~\cite{frome2013devise,kiela2014learning,both2017cktitweoablao,thoma2017knowledge} for capturing visual representation of lexical items. There are various ways of combining multiple representations. Some of the works extract lists of relations from knowledge bases and use those to either modify the learning algorithms~\cite{halawi2012large,wang2014knowledge,Tian2016,rastogi2015multiview} or post-process pre-trained word representations~\cite{faruqui:2014:NIPS-DLRLW}. Another line of literature prepares dense vector representation from each of the modes (text, knowledge bases, visual etc.) and tries to combine the vectors using various methods like concatenation, centroid computation, principal component analysis~\cite{jolliffe1986principal}, canonical correlation analysis~\cite{faruqui2014improving} etc. One such recent attempt is made by~\citet{goikoetxea2016single} where they prepare vector representation from WordNet following the method proposed by~\citet{goikoetxea2015random}, which combines random walks over knowledge bases and neural network language model, and tries to improve the vector representation constructed from text using this. As in lexical knowledge bases, the number of lexical items involved is much less than the raw text and preparing such resources is a cumbersome task, our investigation tries to find whether we can use DT network instead of some knowledge bases like WordNet and achieve comparable performance on NLP tasks like word similarity and word relatedness. 
In order to prepare vector representation from DT network, we attempt to use various network embeddings like DeepWalk~\cite{Perozzi:2014:DOL:2623330.2623732}, LINE~\cite{tang2015line}, struc2vec~\cite{ribeiro2017struc2vec}, node2vec~\cite{grover2016node2vec} etc. Some of those try to capture the neighbourhood or community structure in the network while some attempt to capture structural similarity between nodes, second order proximity, etc. %In our study, node2vec proves to be the best network embedding methods for DT network when tested on word similarity and word relatedness tasks. 

\section{Proposed Methodology}
Our aim is to analyze the effect of integrating the knowledge of Distributional Thesaurus network with the state-of-the-art word representation models to prepare a better word representation. We first prepare vector representations from Distributional Thesaurus (DT) network applying network representation learning model. Next we combine this thesaurus embedding with state-of-the-art vector representations prepared using GloVe and Word2vec model for analysis.
%\vspace{-0.2cm}
\subsection{Distributional Thesaurus (DT) Network}
\begin{figure}[!h]
\begin{center}
%\fbox{\parbox{6cm}{
%This is a figure with a caption.}}
\includegraphics[scale=0.62]{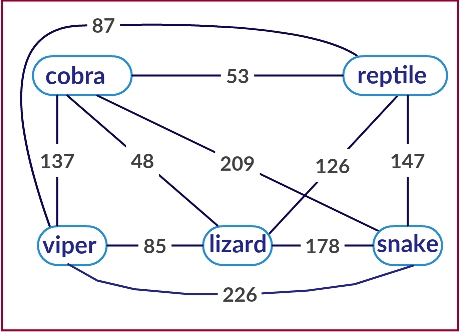} 
\vspace{-0.2cm}
\caption{\textbf{A sample snapshot of Distributional Thesaurus network where each node represents a word and the weight of an edge between two words is defined as the number of context features that these two words share in common.}}
% Here, for a target word `cobra', its hypernym `reptile', co-hyponym `lizard' all are present in its neighborhood.}}
\label{DT}
\end{center}
\vspace{-0.5cm}
\end{figure}

\citet{riedl2013scaling} use the Google books corpus, consisting of texts from over 3.4 million digitized English books published between 1520 and 2008 and construct a distributional thesauri (DT) network using the syntactic n-gram data~\cite{goldberg2013dataset}.
The authors first compute the lexicographer's mutual information (LMI)~\cite{kilgarriff2004itri}\iffalse \footnote{$$
LMI(w,f)=F(w,f)*\log_2 (\nicefrac{F(w,f)}{F(w)*F(f)})$$, where $F()$ measures the frequency, w is word, f is feature}\fi
for each bigram, which gives a measure of the collocational strength of a bigram. Each bigram is broken into a word and a feature, where the feature consists of the bigram relation and the related word. Then the top 1000 ranked features for each word are taken and for each word
pair, intersection of their corresponding feature set is obtained. The word pairs having number of overlapping features above a threshold are retained in the network. In a nutshell, the DT network contains, for each word, a list of words that are similar with respect to their bigram distribution~\cite{riedl2013scaling}. In the network, each word is a node and there is a weighted edge between a pair of words where the weight corresponds to the number of overlapping features. A sample snapshot of the DT is shown in Figure~\ref{DT}.
\vspace{-0.2cm}

\subsection{Embedding Distributional Thesaurus}
Now, from the DT network, we prepare the vector representation for each node using network representation learning models which produce vector representation for each of the node in a network. For this purpose, we use three state-of-the-art network representation learning models as discussed below.

\noindent{\bf{DeepWalk:}} DeepWalk~\cite{Perozzi:2014:DOL:2623330.2623732} learns social representations of a graph's vertices by modeling a stream of short random walks. Social representations signify latent features of the vertices that capture neighborhood similarity and community membership.\\ 
\noindent{\bf{LINE:}} LINE~\cite{tang2015line} is a network embedding model suitable for arbitrary types of networks: undirected, directed and/or weighted. The model optimizes an objective which preserves both the local and global network structures by capturing both first-order and second-order proximity between vertices.\\ 
\noindent{\bf{node2vec:}} node2vec~\cite{grover2016node2vec} is a semi-supervised algorithm for scalable feature learning in networks which maximizes the likelihood of preserving network neighborhoods of nodes in a d-dimensional feature space. This algorithm can learn representations that organize nodes based on their network roles and/or communities they belong to by developing a family of biased random walks, which efficiently explore diverse neighborhoods of a given node.\\
Note that, by applying network embedding models on DT network we obtain 128 dimensional vector for each word in the network. We only consider edges of the DT network having edge weight greater or equal to 50 for network embedding.
Henceforth, we will use {\sl D2V-D}, {\sl D2V-L} and {\sl D2V-N} to indicate vector representations obtained from DT network produced by DeepWalk, LINE and node2vec, respectively.
\if{0}
\subsection{GloVe} As proposed by~\citet{pennington2014glove}, GloVe is a count based unsupervised learning algorithm for obtaining vector representations for words. Authors use a  specific weighted least squares model that trains on global word-word co-occurrence counts and thus makes efficient use of statistics. The resulting vector representation is proved to be successfully showcasing interesting linear substructures of the word vector space. A very well-known GloVe 1.2 embeddings trained on 840 billion words of the common crawl dataset having vector dimension of 300 is publicly available. We directly use this pre-trained GloVe vectors for comparison. 
\subsection{Word2vec} Word2vec is a group of predictive models used to produce word embeddings as introduced by~\citet{mikolov2013efficient}. Two-layer neural networks are trained to reconstruct linguistic contexts of words and resulting word vectors are positioned in the vector space such that words that share common contexts in the corpus are located in close proximity to one another in the space. We use the prominent pre-trained vector representations prepared by the authors trained on 100 billion words of Google News using skip-grams with negative sampling having dimension of 300. For our analysis we download this representation and use as it is.   
\fi

After obtaining vector representations, we also explore whether these can be combined with the pre-trained vector representation of Word2vec and GloVE to come up with a joint vector representation. For that purpose, we directly use very well-known GloVe 1.2 embeddings~\cite{pennington2014glove} trained on 840 billion words of the common crawl dataset having vector dimension of 300. As an instance of pre-trained vector of Word2vec, we use prominent pre-trained vector representations prepared by~\citet{mikolov2013efficient} trained on 100 billion words of Google News using skip-grams with negative sampling, having dimension of 300.
%\vspace{-0.2cm}
\subsection{Vector Combination Methods}
%\vspace{-0.2cm}
In order to integrate the word vectors, we apply two strategies inspired by ~\citet{goikoetxea2016single}.\\
\noindent{\bf{Concatenation (CC):}} This corresponds to the simple vector concatenation operation. Vector representations of both GloVe and Word2vec are of 300 dimensions and word embeddings learnt form DT are of 128 dimensions. The concatenated representation we use are of 428 dimensions.\\   
\noindent{\bf{Principal Component Analysis (PCA):}} Principal component analysis~\cite{jolliffe1986principal} is a dimensionality reduction statistical procedure that uses an orthogonal transformation to convert a set of observations of possibly correlated variables into a set of values of linearly uncorrelated variables called principal components (linear combinations of the original variables). We apply PCA to the concatenated representations (dimension of 428) reducing these to 300 dimensions. 
In addition to PCA, we try with truncated singular value decomposition procedure~\cite{hansen1987truncatedsvd} as well, but as per the experiment set up, it shows negligible improvement in performance compared to simple concatenation; hence we do not continue with the truncated singular value decomposition for dimensionality reduction. After obtaining the combined representations of words, we head towards evaluating the quality of the representation.

\section{Experiments and Analysis}
\begin{table*}[!tbh]
\begin{center}
\begin{small}
    \begin{tabular}{|>{\centering}p{2cm}|>{\centering}p{1.5cm}|>{\centering}p{1.5cm}|>{\centering}p{1.5cm} >{\centering}p{1.5cm}>{\centering}p{1.5cm}|}
    \hline
\bf{Dataset} &	\bf{GloVe} & \bf{W2V} & \bf{D2V-D} & \bf{D2V-L} & \bf{D2V-N} \tabularnewline \hline
WSSim &	\bf{0.799}	& 0.779 &  0.737	& 0.073	&	0.764 \tabularnewline %\hline 
SimL-N 	& 0.427 &	\bf{0.454} & 0.418	& 0.015 & 0.421 \tabularnewline %\hline 
RG-65 	& 0.791 & 0.777 & 0.804 & -0.121 & \bf{0.813} \tabularnewline %\hline 
MC-30 	& 0.799 &	0.819 & 0.859	& -0.067 & \bf{0.869} \tabularnewline \hline\hline 
WSR 	& \bf{0.637} & 0.631 & 0.287 & 0.077 & 0.333 \tabularnewline %\hline 
M771 	& \bf{0.707} &	0.655 & 0.636	& 0.027 & 0.63 \tabularnewline %\hline
M287 	& \bf{0.8} &	0.755 & 0.558	& -0.027 & 0.591 \tabularnewline %\hline
MEN-N 	& \bf{0.819} &	0.764 & 0.619	& 0.004 & 0.612 \tabularnewline \hline \hline 
WS-353 	& \bf{0.706} & 0.697 & 0.51	& 0.088 & 0.547 \tabularnewline \hline 
\end{tabular}
\end{small}
\end{center}

%\vspace{-0.3cm}
\caption{\textbf{Comparison of individual performances of different vector representation models w.r.t. word similarity and relatedness tasks. The performance metric is Spearman's rank correlation coefficient ($\rho$). Best result of each row in bold showing the best vector representation for each dataset.}}
\label{exp1}
%\vspace{-0.1cm}
\end{table*}

In order to evaluate the quality of the word representations, we follow the most acceptable way of applying on different NLP tasks like word similarity and word relatedness, synonym detection and word analogy as described next. 
%\vspace{-0.2cm}
\subsection{Word Similarity and Relatedness}
In this genre of tasks, the human judgment score for each word pair is given; we report the Spearman's rank correlation coefficient ($\rho$) between human judgment score and the predicted score by distributional model. Note that, we take cosine similarity between vector representations of words in a word pair as the predicted score.

\begin{table*}[!tbh]
\begin{center}
\begin{small}
    \begin{tabular}{|>{\centering}p{1.4cm}|>{\centering}p{0.9cm}|>{\centering}p{2.5cm}>{\centering}p{2.5cm}|>{\centering}p{2.5cm}>{\centering}p{2.5cm}|}
    \hline
\bf{Dataset} &	\bf{GloVe} & \bf{CC (GloVe,D2V-D)} & \bf{PCA (GloVe,D2V-D)} & \bf{CC (GloVe,D2V-N)} & \bf{PCA (GloVe,D2V-N)} \tabularnewline \hline
WSSim &	0.799	& 0.838 &  0.839	& \bf{0.84}	&	0.832 \tabularnewline %\hline 
SimL-N 	& 0.427 &	0.443 & 0.468	& 0.446 & \bf{0.483} \tabularnewline %\hline 
RG-65 	& 0.791 & 0.816 & \bf{0.879} & 0.809 & 0.857 \tabularnewline %\hline 
MC-30 	& 0.799 &	0.86 & \bf{0.89} & 0.866 & 0.874 \tabularnewline \hline\hline 
WSR 	& 0.637 & \bf{0.676} & 0.645 & 0.67 & 0.657 \tabularnewline %\hline 
M771 	& 0.707 &	0.708 & 0.707	& 0.711 & \bf{0.719} \tabularnewline %\hline 
M287 	& 0.8 &	0.781 & 0.807	& 0.795 & \bf{0.82} \tabularnewline %\hline 
MEN-N 	& \bf{0.819} & 0.792 & 0.799	& 0.806 & 0.817 \tabularnewline \hline \hline 
WS-353 	& 0.706 & \bf{0.751} & 0.74	& 0.75 & 0.75 \tabularnewline \hline 
\end{tabular}
\end{small}
\end{center}
%\vspace{-0.4cm}
\caption{\textbf{ Comparison of performances (Spearman's $\rho$) of GloVe against the combined representation of word representations obtained from DT network using network embeddings (DeepWalk, node2vec) with GloVe. Two combination methods -- Concatenation (CC) and PCA -- are used among which PCA performs better than concatenation (CC) in most of the cases. Also the result shows that the combined representation leads to better performance in almost all the cases.}}
\label{tab:ex2}
\end{table*}

\begin{table*}[!tbh]
\begin{center}
\begin{small}
    \begin{tabular}{|>{\centering}p{1.4cm}|>{\centering}p{0.9cm}|>{\centering}p{2.5cm}>{\centering}p{2.5cm}|>{\centering}p{2.5cm}>{\centering}p{2.5cm}|}
    \hline
\bf{Dataset} &	\bf{W2V} & \bf{CC (W2V,D2V-D)} & \bf{PCA (W2V,D2V-D)} & \bf{CC (W2V,D2V-N)} & \bf{PCA (W2V,D2V-N)} \tabularnewline \hline
WSSim &	0.779	& 0.774 &  0.786	& \bf{0.806}	&	0.805 \tabularnewline %\hline 
SimL-N 	& 0.454 &	0.438 & 0.456	& 0.448 & \bf{0.493} \tabularnewline %\hline 
RG-65 	& 0.777 & 0.855 & 0.864 & 0.867 & \bf{0.875} \tabularnewline %\hline 
MC-30 	& 0.819 &	0.866 & 0.891 & 0.903 & \bf{0.909} \tabularnewline \hline\hline 
WSR 	& \bf{0.631} & 0.441 & 0.443 & 0.459 & 0.497 \tabularnewline %\hline 
M771 	& 0.655 &	0.633 & 0.637	& 0.656 & \bf{0.676} \tabularnewline %\hline 
M287 	& \bf{0.755} &	0.714 & 0.701	& 0.722 & \bf{0.755} \tabularnewline %\hline 
MEN-N 	& \bf{0.764} & 0.703 & 0.717	& 0.714 & 0.747 \tabularnewline \hline \hline 
WS-353 	& \bf{0.697} & 0.602 & 0.61	& 0.623 & 0.641 \tabularnewline \hline 
\end{tabular}
\end{small}
\end{center}
%\vspace{-0.2cm}
\caption{\textbf{A similar experiment as Table~\ref{tab:ex2} with Word2vec (W2V) instead of GloVe.}}
\label{tab:ex3}
%\vspace{-0.2cm}
\end{table*}

\begin{table}[!h]
\centering
\begin{small}
\begin{tabular}{|>{\centering}p{1.4cm}|>{\centering}p{2.5cm}>{\centering}p{2.5cm}|}
    \hline
\bf{Dataset} &	\bf{PCA (GloVe,W2V)} & \bf{PCA (GloVe,D2V-N)} \tabularnewline \hline
WSSim &	0.8	& \bf{0.832} \tabularnewline %\hline 
SimL-N 	& 0.476 & \bf{0.483}  \tabularnewline %\hline 
RG-65 	& 0.794 & \bf{0.857}  \tabularnewline %\hline 
MC-30 	& 0.832 &	\bf{0.874}  \tabularnewline \hline\hline 
WSR 	& \bf{0.68} & 0.657  \tabularnewline %\hline 
M771 	& 0.717 &	\bf{0.719}  \tabularnewline %\hline 
M287 	& \bf{0.82} &	\bf{0.82}  \tabularnewline %\hline 
MEN-N 	& \bf{0.829} & 0.817  \tabularnewline \hline \hline 
WS-353 	& 0.746 & \bf{0.75}  \tabularnewline \hline 
\end{tabular}
\end{small}
%\vspace{-0.2cm}
\caption{\bf{Comparison of performances (Spearman's $\rho$) between GloVe combined with Word2vec (W2V) against GloVe combined with DT embedding obtained using node2vec (D2V-N). PCA has been taken as combination method. Clearly, DT embedding outperforms Word2vec in terms of enhancing the performance of GloVe.}}
\label{tab:ex4}
% \end{center}
%\vspace{-0.5cm}
\end{table}
\if{0}
\begin{table*}[!tbh]
\begin{center}
\begin{small}
    \begin{tabular}{|>{\centering}p{1.4cm}|>{\centering}p{2.5cm}>{\centering}p{2.5cm}>{\centering}p{4cm}|}
    \hline
\bf{Dataset}&	\bf{PCA (GloVe,D2V-N)} & \bf{PCA (GloVe,WN2V)} & \bf{PCA (GloVe,WN2V,D2V-N)} \tabularnewline \hline
WSSim &	0.832	& 0.828 & \bf{0.853}  \tabularnewline %\hline 
SimL-N 	& 0.483 & 0.525 & \bf{0.531} \tabularnewline %\hline 
RG-65 	& 0.857 & 0.858 & \bf{0.91} \tabularnewline %\hline 
MC-30 	& 0.874 & 0.882 & \bf{0.92} \tabularnewline \hline\hline 
WSR 	& 0.657 & \bf{0.699} & 0.682  \tabularnewline %\hline 
M771 	& 0.719 & 0.762 & \bf{0.764} \tabularnewline %\hline 
M287 	& \bf{0.82} & 0.816 & 0.81 \tabularnewline %\hline 
MEN-N 	& 0.817 & \bf{0.848} & 0.7993  \tabularnewline \hline \hline 
WS-353 	& 0.75 & \bf{0.7801} & 0.7693 \tabularnewline \hline 
\end{tabular}
\end{small}
\end{center}

\vspace{-0.2cm}
\caption{\textbf{Spearman's Rank Correlation coefficient ($\rho$) reported for combined representation of GloVe and DT embedding using node2vec (D2V-N), combined representation of GloVe and WordNet embedding (WN2V), combined representation of GloVe, WN2V and D2V-N. Results show that, DT embedding produces comparable performance w.r.t WordNet embedding even if it is not a manually created resource. Also combining DT embedding along with WordNet embedding helps to boost performance further in many of the cases.}}
\label{tab:ex5}
%\vspace{-0.2cm}
\end{table*}
\fi

\begin{table}[!h]
\centering
\begin{small}
\begin{tabular}{|>{\centering}p{1.1cm}|>{\centering}p{1.1cm}>{\centering}p{1.8cm}>{\centering}p{2.1cm}|}
    \hline
\bf{Dataset}&	\bf{PCA (GloVe, D2V-N)} & \bf{PCA (GloVe, WN2V)} & \bf{PCA (GloVe, WN2V, D2V-N)}\tabularnewline \hline
WSSim &	0.832	& 0.828 & \bf{0.853}  \tabularnewline %\hline 
SimL-N 	& 0.483 & 0.525 & \bf{0.531} \tabularnewline %\hline 
RG-65 	& 0.857 & 0.858 & \bf{0.91} \tabularnewline %\hline 
MC-30 	& 0.874 & 0.882 & \bf{0.92} \tabularnewline \hline\hline 
WSR 	& 0.657 & \bf{0.699} & 0.682  \tabularnewline %\hline 
M771 	& 0.719 & 0.762 & \bf{0.764} \tabularnewline %\hline 
M287 	& \bf{0.82} & 0.816 & 0.81 \tabularnewline %\hline 
MEN-N 	& 0.817 & \bf{0.848} & 0.7993  \tabularnewline \hline \hline 
WS-353 	& 0.75 & \bf{0.7801} & 0.7693 \tabularnewline \hline 
\end{tabular}
\end{small}
\vspace{-0.2cm}
\caption{\textbf{ Performance ($\rho$) reported for three combined representations: GloVe and DT embedding using node2vec (D2V-N), GloVe and WordNet embedding (WN2V), GloVe, WN2V and D2V-N. Results show that, DT embedding produces comparable performance w.r.t WordNet embedding. Combining DT embedding along with WordNet embedding helps to boost performance in further many of the cases.}}
\label{tab:exp5}
%\vspace{-0.1cm}
% \end{center}
\end{table}

\begin{table}[!h]
\centering
\begin{small}
\begin{tabular}{|>{\centering}p{1.1cm}|>{\centering}p{0.9cm}|>{\centering}p{1.8cm}>{\centering}p{2.1cm}|}
    \hline
\bf{Dataset} &	\bf{GloVe} & \bf{GloVe with retrofitting} & \bf{PCA (GloVe,D2V-N)} \tabularnewline \hline
WSSim &	0.799 & 0.799 & \bf{0.832} \tabularnewline %\hline 
SimL-N 	& 0.427 & 0.423 & \bf{0.483}  \tabularnewline %\hline 
RG-65 	& 0.791 & 0.791 & \bf{0.857}  \tabularnewline %\hline 
MC-30 	& 0.799 & 0.799 &\bf{0.874} \tabularnewline \hline\hline 
WSR 	& 0.637 & \bf{0.69} & 0.657 \tabularnewline %\hline 
M771 	& 0.707 & 0.708 & \bf{0.719}  \tabularnewline %\hline
M287 	& 0.8 & 0.795 & \bf{0.82}  \tabularnewline %\hline
MEN-N 	& \bf{0.819} & \bf{0.819} & 0.817  \tabularnewline \hline \hline 
WS-353 	& 0.706 & 0.703 & \bf{0.75}  \tabularnewline \hline 
\end{tabular}
\end{small}
\vspace{-0.2cm}
\caption{\bf{Comparison of performances (Spearman's $\rho$) between GloVe representation and retrofitted (by DT network) GloVe representation. Clearly, DT retrofitting is not helping much to improve the performance of GloVe.}}
\label{tab:exp6}
%\vspace{-0.2cm}
% \end{center}
\end{table}
%\fi
\noindent{\bf{Datasets:}}
We use the benchmark datasets for evaluation of word representations. Four word similarity datasets and four word relatedness datasets are used for that purpose. The descriptions of the word similarity datasets are given below.\\
%\begin{itemize}[leftmargin=*]
%\item 
\noindent{\textbf{WordSim353 Similarity (WSSim) :}} 203 word pairs extracted from WordSim353 dataset~\cite{finkelstein2001placing} by manual classification, prepared by~\citet{agirre2009study}, which deals with only similarity.\\ 
\noindent{ \textbf{SimLex999 (SimL) :}} 999 word pairs rated by 500 paid native English speakers, recruited via Amazon Mechanical Turk,\footnote{\url{www.mturk.com}} who were asked to rate the similarity. This dataset is introduced by~\citet{hill2016simlex}.\\
\noindent{\textbf{RG-65 :}} It consists of 65 word pairs collected by~\citet{rubenstein1965contextual}. These word pairs are judged by 51 humans in a scale from 0 to 4 according to their similarity, but ignoring any other possible semantic relationships.\\
\noindent{\textbf{MC-30 :}} 30 words judged by 38 subjects in a scale of 0 and 4 collected by~\citet{miller1991contextual}.

Similarly, a brief overview of word relatedness datasets is given below:\\
%\begin{itemize}[leftmargin=*]
\noindent{ \textbf{WordSim353 Relatedness (WSR) :}} 252 word pairs extracted from WordSim353~\cite{finkelstein2001placing} dataset by manual classification, prepared by~\citet{agirre2009study} which deals with only relatedness.\\
\noindent{\textbf{MTURK771 (M771) :}} 771 word pairs evaluated by Amazon Mechanical Turk
workers, with an average of 20 ratings for each word pair, where each judgment task consists of a batch of 50 word pairs. Ratings are collected on a 1–5 scale. This dataset is introduced by~\citet{halawi2012large}.\\
\noindent{\textbf{MTURK287 (M287) :}} 287 word pairs evaluated by Amazon Mechanical Turk
workers, with an average of 23 ratings for each word pair. This dataset is introduced by~\citet{radinsky2011word}.\\
\noindent{\textbf{MEN :}} MEN consists of 3,000 word pairs with [0, 1]-normalized semantic relatedness ratings provided by Amazon Mechanical Turk workers. This dataset was introduced by~\citet{bruni2014multimodal}.

Along with these datasets we use the full \textbf{WordSim353 (WS-353)} dataset (includes both similarity and relatedness pairs)~\cite{finkelstein2001placing} which contains 353 word pairs, each associated with an average of 13 to 16 human judgments in a scale of 0 to 10. Being inspired by~\citet{baroni2014don}, we consider only noun pairs from \textbf{SimL} and \textbf{MEN} datasets, which will be denoted as \textbf{SimL-N} and \textbf{MEN-N} whereas other datasets contain noun pair only. 
%We are report the Spearman's rank correlation ($\rho$) between human judgment score and cosine similarity score of vector representations of word pair.
%we perform Fisher's Z-transformation~\cite{press2007numerical} for significance testing.

We start with experiments to inspect individual performance of each of the vector representations for each of the datasets. Table~\ref{exp1} represents individual performances of GloVe, Word2vec, {\sl D2V-D, D2V-L} and {\sl D2V-N} for different datasets. In most of the cases, GloVe produces the best results although no model is a clear winner for all the datasets. Interestingly, {\sl D2V-D} and {\sl D2V-N} give results comparable to GloVe and Word2vec for the word similarity datasets, even surpassing GloVe and Word2vec for few of these. {\sl D2V-L} gives very poor performance, indicating that considering second order proximity in the DT network while embedding has an adverse effect on performance in word similarity and word relatedness tasks, whereas random walk based {\sl D2V-D} and {\sl D2V-N} which take care of neighborhood and community, produce decent performance. Henceforth, we ignore the {\sl D2V-L} model for the rest of our experiments.

Next, we investigate whether network embeddings applied on Distributional Thesaurus network can be combined with GloVe and Word2vec to improve  the performance on the pre-specified tasks. In order to do that, we combine the vector representations using two operations: concatenation (CC), and principal component analysis (PCA). Table~\ref{tab:ex2} represents the performance of combining GloVe with {\sl D2V-D} and {\sl D2V-N} for all the datasets using these combination strategies. In general, PCA turns out to be better technique for vector combination than CC. Clearly, combining DT embeddings and GloVe boosts the performance for all the datasets except for the \textbf{MEN-N} dataset, where the combined representation produces comparable performance.

In order to ensure that this observation is consistent, we try combining DT embeddings with Word2vec. The results are presented in Table~\ref{tab:ex3} and we see the same improvement in the performance except for a few cases, indicating the fact that combining word embeddings prepared form DT network is helpful in enhancing performances. From Tables~\ref{exp1},~\ref{tab:ex2} and \ref{tab:ex3} we see that GloVe proves to be best for most of the cases, {\sl D2V-N} is the best performing network embedding, and PCA turns out to be the best combination technique. Henceforth, we consider PCA (GloVe, {\sl D2V-N}) as our model for comparison with the baselines for the rest of the experiments.

Further, to scrutinize that the achieved result is not just the effect of combining two different word vectors, we compare PCA (GloVe, D2V-N) against combination of GloVe and Word2vec (W2V). Table ~\ref{tab:ex4} shows the performance comparison on different datasets and it is evident that PCA (GloVe, D2V-N) gives better results compared to PCA (GloVe, W2V) in most of the cases.

Now, as we observe that the network embedding from DT network helps to boost the performance of Word2vec and GloVe when combined with them, we further compare the performance against the case when text based embeddings are combined with embeddings from lexical resources. For that purpose, we take \textbf{one baseline}~\cite{goikoetxea2016single}, where authors combined the text based representation with WordNet based representation. Here we use GloVe as the text based representation and PCA as the combination method as prescribed by the author. Note that, WordNet based representation is made publicly available by~\citet{goikoetxea2016single}. From the second and third columns of Table~\ref{tab:exp5}, we observe that even though we do not use any manually created lexical resources like WordNet our approach achieves comparable performance. Additionally we check whether we gain in terms of performance if we integrate the three embeddings together. Fourth column of Table~\ref{tab:exp5} shows that we gain for some of the datasets and for other cases, it has a negative effect. Looking at the performance, we can conclude that automatically generated DT network from corpus brings in useful additional information as far as word similarity and relatedness tasks are concerned.

So far, we use concatenation and PCA as methods for combining two different representations. However, as per the literature, there are different ways of infusing knowledge from different lexical sources to improve the quality of pre-trained vector embeddings. So we compare our proposed way of combination with a completely different way of integrating information from both dimensions, known as {\sl retrofitting}. Retrofitting is a novel way proposed by~\citet{faruqui:2014:NIPS-DLRLW} for refining vector space representations using relational information from semantic lexicons by encouraging linked
words to have similar vector representations. Here instead of using semantic lexicons, we use the DT network to produce the linked words to have similar vector representation. Note that, for a target word, we consider only those words as linked words which are having edge weight greater than a certain threshold. While experimenting with various thresholds, the best results were obtained for a threshold value of 500. Table~\ref{tab:exp6} shows the performance of GloVe representations when retrofitted with information from DT network. Even though in very few cases it gives little improved performance, compared to other combinations presented in Table\ref{tab:ex2}, the correlation is not very good, indicating the fact that retrofitting is probably not the best way of fusing knowledge from a DT network.

Further, we extend our study to investigate the usefulness of DT embedding on other NLP tasks like synonym detection, SAT analogy task as will be discussed next.
%\vspace{-0.2cm}
\subsection{Synonym Detection}
We consider two gold standard datasets for the experiment of synonym detection. The descriptions of the used datasets are given below.\\ 
\noindent{\textbf{TOEFL: }}It contains 80 multiple-choice synonym questions (4 choices per question) introduced by~\citet{landauer1997solution}, as a way of evaluating algorithms for measuring degree of similarity between words. Being consistent with the previous experiments, we consider only nouns for our experiment and prepare \textbf{TOEFL-N} which contains 23 synonym questions. \\ 
\noindent{\textbf{ESL: }}It contains 50 multiple-choice synonym questions (4 choices per question), along with a sentence for providing context for each of the question, introduced by~\citet{turney2001mining}. Here also we consider only nouns for our experiment and prepare \textbf{ESL-N} which contains 22 synonym questions. Note that, in our experimental setup we do not use the context per question provided in the dataset for evaluation.\\
While preparing both the datasets we also keep in mind the availability of word vectors in both downloaded GloVe representation and prepared DT embedding.
For evaluation of the word embeddings using \textbf{TOEFL-N} and \textbf{ESL-N}, we consider the option as the correct answer which is having highest cosine similarity with the question and report accuracy. From the results presented in Table~\ref{tab:exp7}, we see that DT embedding leads to boost the performance of GloVe representation.     
\begin{table}[!h]
%\vspace{-0.4cm}
\centering
\begin{small}
\begin{tabular}{|>{\centering}p{1.7cm}|>{\centering}p{0.9cm}>{\centering}p{1cm}>{\centering}p{2.5cm}|}
    \hline
\bf{Dataset} &	\bf{GloVe} & \bf{D2V-N} & \bf{PCA (GloVe, D2V-N)} \tabularnewline \hline
TOEFL-N & 0.826 & 0.739 & \bf{0.869}\tabularnewline %\hline 
ESL-N 	& 0.636 & 0.591 &\bf{0.682}  \tabularnewline \hline\hline
SAT-N &	0.465 & 0.509 & \textbf{0.515} 
\tabularnewline \hline
\end{tabular}
\end{small}
%\vspace{-0.4cm}
\caption{\bf{Comparison of accuracies between GloVe representation, DT embedding using node2vec and combination of both where PCA is the combination technique. Clearly DT embedding is helping to improve the performance of GloVe for synonym detection as well as analogy detection.}}
\label{tab:exp7}
% \end{center}
%\vspace{-0.2cm}
\end{table}
%\vspace{-0.2cm}
\subsection{Analogy Detection}
For analogy detection we experiment with \textbf{SAT} analogy dataset. This dataset contains 374 multiple-choice analogy questions (5 choices per question) introduced by~\citet{turney2003combining} as a way of evaluating algorithms for measuring relational similarity. Considering only noun questions, we prepare \textbf{SAT-N}, which contains 159 questions.\\ 
In order to find out the correct answer from the 5 options given for each question, we take up a score ($s$) metric proposed by~\citet{speer2017conceptnet}, where for a question `$a_1$ is to $b_1$', we will consider `$a_2$ is to $b_2$' as the correct answer among the options, whose score ($s$) is the highest. Score ($s$) is defined by the author as follows:

%\begin{equation*}
%\begin{multlined}
$s=a_1.a_2 + b_1.b_2 + w_1(b_2-a_2).(b_1-a_1) + w_2(b_2-b_1).(a_2-a_1)$\\
%\end{multlined}
%\end{equation*}
As mentioned by the authors, the appropriate values of $w_1$ and $w_2$ is optimized separately for each system using grid search, to achieve the best performance. We use accuracy as the evaluation metric. The last row of Table~\ref{tab:exp7} presents the comparison of accuracies (best for each model) obtained using different embeddings portraying the same observation that combination of GloVe and DT embeddings leads to better performance compared to GloVe and DT embeddings when used separately.
%as well as the combination of both embeddings.
%which comes up with little off the track observation, portraying DT embedding as a clear winner compared to GloVe as well as the combination of both embeddings. Even though it is difficult to come up with a conclusion after trying with single dataset for analogy task, the performance of DT embedding looks promising for future investigation. 
Note that, the optimized values of $(w_1,w_2)$ are (0.2,0.2), (0.8,0.6), (6,0.6) for GloVe, DT embedding, combined representation of GloVe and DT embeddings respectively, for the analogy task.   
\if{0}
\begin{table}[!h]
\centering
\begin{tabular}{|>{\centering}p{1.3cm}|>{\centering}p{0.9cm}|>{\centering}p{1.2cm}|>{\centering}p{2.5cm}|}
    \hline
\bf{Dataset} &	\bf{GloVe} & \bf{D2V-N} & \bf{PCA (GloVe,D2V-N)} \tabularnewline \hline
SAT-N &	0.465 & \textbf{0.509} & {0.465}  \tabularnewline \hline 
\end{tabular}
\caption{\bf{Comparison of performances (Accuracy) of GloVe representation, DT embedding using node2vec and combination of both where PCA is the combination technique. Even though combination of both representation doesn't lead to better performance, individual performance of DT embedding is quite promising.}}
\label{tab:exp8}
% \end{center}
\end{table}
\fi

\section{Conclusion}
In this paper we showed that both dense count based model (GloVe) and predictive model (Word2vec) lead to improved word representation when they are combined with word representation learned using network embedding methods on Distributional Thesaurus (DT) network. We tried with various network embedding models among which {\sl node2vec} proved to be the best in our experimental setup. We also tried with different methodologies to combine vector representations and PCA turned out to be the best among them. The combined vector representation of words yielded the better performance for most of the similarity and relatedness datasets as compared to the performance of GloVe and Word2vec representation individually. Further we observed that we could use the information from DT as a proxy of WordNet in order to improve the state-of-the-art vector representation as we were getting comparable performances for most of the datasets. Similarly, for synonym detection task and analogy detection task, the same trend of combined vector representation continued, showing the superiority of the combined representation over state-of-the-art embeddings.
%whereas for analogy task DT embedding outperformed both GloVe and the combined representation. 
All the datasets used in our experiments which are not under any copyright protection, along with the DT embeddings are made publicly available\footnote{\url{https://tinyurl.com/yamrutrm}}.

In future we plan to investigate the effectiveness of the joint representation on other NLP tasks like text classification, sentence completion challenge, evaluation of common sense stories etc. The overall aim is to prepare a better generalized representation of words which can be used across languages in different NLP tasks.

\if{0}
\section{Introduction}

The following instructions are directed to authors of papers submitted
to NAACL-HLT 2018 or accepted for publication in its proceedings. All
authors are required to adhere to these specifications. Authors are
required to provide a Portable Document Format (PDF) version of their
papers. \textbf{The proceedings are designed for printing on A4
paper.}
\fi

%\section{General Instructions}
\if{0}
Manuscripts must be in two-column format.  Exceptions to the
two-column format include the title, authors' names and complete
addresses, which must be centered at the top of the first page, and
any full-width figures or tables (see the guidelines in
Subsection~\ref{ssec:first}). {\bf Type single-spaced.}  Start all
pages directly under the top margin. See the guidelines later
regarding formatting the first page.  The manuscript should be
printed single-sided and its length
should not exceed the maximum page limit described in Section~\ref{sec:length}.
Pages are numbered for  initial submission. However, {\bf do not number the pages in the camera-ready version}.

By uncommenting {\small\verb|\aclfinalcopy|} at the top of this 
 document, it will compile to produce an example of the camera-ready formatting; by leaving it commented out, the document will be anonymized for initial submission.  When you first create your submission on softconf, please fill in your submitted paper ID where {\small\verb|***|} appears in the {\small\verb|\def\aclpaperid{***}|} definition at the top.

The review process is double-blind, so do not include any author information (names, addresses) when submitting a paper for review.  
However, you should maintain space for names and addresses so that they will fit in the final (accepted) version.  The NAACL-HLT 2018 \LaTeX\ style will create a titlebox space of 2.5in for you when {\small\verb|\aclfinalcopy|} is commented out.  

The author list for submissions should include all (and only) individuals who made substantial contributions to the work presented. Each author listed on a submission to NAACL-HLT 2018 will be notified of submissions, revisions and the final decision. No authors may be added to or removed from submissions to NAACL-HLT 2018 after the submission deadline.

\subsection{The Ruler}
The NAACL-HLT 2018 style defines a printed ruler which should be presented in the
version submitted for review.  The ruler is provided in order that
reviewers may comment on particular lines in the paper without
circumlocution.  If you are preparing a document without the provided
style files, please arrange for an equivalent ruler to
appear on the final output pages.  The presence or absence of the ruler
should not change the appearance of any other content on the page.  The
camera ready copy should not contain a ruler. (\LaTeX\ users may uncomment the {\small\verb|\aclfinalcopy|} command in the document preamble.)  
Reviewers: note that the ruler measurements do not align well with
lines in the paper -- this turns out to be very difficult to do well
when the paper contains many figures and equations, and, when done,
looks ugly. In most cases one would expect that the approximate
location will be adequate, although you can also use fractional
references ({\em e.g.}, the first paragraph on this page ends at mark $108.5$).

\subsection{Electronically-available resources}

NAACL-HLT provides this description in \LaTeX2e{} ({\small\tt naaclhlt2018.tex}) and PDF
format ({\small\tt naaclhlt2018.pdf}), along with the \LaTeX2e{} style file used to
format it ({\small\tt naaclhlt2018.sty}) and an ACL bibliography style ({\small\tt acl\_natbib.bst})
and example bibliography ({\small\tt naaclhlt2018.bib}).
These files are all available at
{\small\tt http://naacl2018.org/downloads/ naaclhlt2018-latex.zip}. 
 We
strongly recommend the use of these style files, which have been
appropriately tailored for the NAACL-HLT 2018 proceedings.

\subsection{Format of Electronic Manuscript}
\label{sect:pdf}

For the production of the electronic manuscript you must use Adobe's
Portable Document Format (PDF). PDF files are usually produced from
\LaTeX\ using the \textit{pdflatex} command. If your version of
\LaTeX\ produces Postscript files, you can convert these into PDF
using \textit{ps2pdf} or \textit{dvipdf}. On Windows, you can also use
Adobe Distiller to generate PDF.

Please make sure that your PDF file includes all the necessary fonts
(especially tree diagrams, symbols, and fonts with Asian
characters). When you print or create the PDF file, there is usually
an option in your printer setup to include none, all or just
non-standard fonts.  Please make sure that you select the option of
including ALL the fonts. \textbf{Before sending it, test your PDF by
  printing it from a computer different from the one where it was
  created.} Moreover, some word processors may generate very large PDF
files, where each page is rendered as an image. Such images may
reproduce poorly. In this case, try alternative ways to obtain the
PDF. One way on some systems is to install a driver for a postscript
printer, send your document to the printer specifying ``Output to a
file'', then convert the file to PDF.

It is of utmost importance to specify the \textbf{A4 format} (21 cm
x 29.7 cm) when formatting the paper. When working with
{\tt dvips}, for instance, one should specify {\tt -t a4}.
Or using the command \verb|\special{papersize=210mm,297mm}| in the latex
preamble (directly below the \verb|\usepackage| commands). Then using 
{\tt dvipdf} and/or {\tt pdflatex} which would make it easier for some.

Print-outs of the PDF file on A4 paper should be identical to the
hardcopy version. If you cannot meet the above requirements about the
production of your electronic submission, please contact the
publication chairs as soon as possible.

\subsection{Layout}
\label{ssec:layout}

Format manuscripts two columns to a page, in the manner these
instructions are formatted. The exact dimensions for a page on A4
paper are:

\begin{itemize}
\item Left and right margins: 2.5 cm
\item Top margin: 2.5 cm
\item Bottom margin: 2.5 cm
\item Column width: 7.7 cm
\item Column height: 24.7 cm
\item Gap between columns: 0.6 cm
\end{itemize}

\noindent Papers should not be submitted on any other paper size.
 If you cannot meet the above requirements about the production of 
 your electronic submission, please contact the publication chairs 
 above as soon as possible.

\subsection{Fonts}

For reasons of uniformity, Adobe's {\bf Times Roman} font should be
used. In \LaTeX2e{} this is accomplished by putting

\begin{quote}
\begin{verbatim}
\usepackage{times}
\usepackage{latexsym}
\end{verbatim}
\end{quote}
in the preamble. If Times Roman is unavailable, use {\bf Computer
  Modern Roman} (\LaTeX2e{}'s default).  Note that the latter is about
  10\% less dense than Adobe's Times Roman font.

\begin{table}[t!]
\begin{center}
\begin{tabular}{|l|rl|}
\hline \bf Type of Text & \bf Font Size & \bf Style \\ \hline
paper title & 15 pt & bold \\
author names & 12 pt & bold \\
author affiliation & 12 pt & \\
the word ``Abstract'' & 12 pt & bold \\
section titles & 12 pt & bold \\
document text & 11 pt  &\\
captions & 10 pt & \\
abstract text & 10 pt & \\
bibliography & 10 pt & \\
footnotes & 9 pt & \\
\hline
\end{tabular}
\end{center}
\caption{\label{font-table} Font guide. }
\end{table}

\subsection{The First Page}
\label{ssec:first}

Center the title, author's name(s) and affiliation(s) across both
columns. Do not use footnotes for affiliations. Do not include the
paper ID number assigned during the submission process. Use the
two-column format only when you begin the abstract.

{\bf Title}: Place the title centered at the top of the first page, in
a 15-point bold font. (For a complete guide to font sizes and styles,
see Table~\ref{font-table}) Long titles should be typed on two lines
without a blank line intervening. Approximately, put the title at 2.5
cm from the top of the page, followed by a blank line, then the
author's names(s), and the affiliation on the following line. Do not
use only initials for given names (middle initials are allowed). Do
not format surnames in all capitals ({\em e.g.}, use ``Mitchell'' not
``MITCHELL'').  Do not format title and section headings in all
capitals as well except for proper names (such as ``BLEU'') that are
conventionally in all capitals.  The affiliation should contain the
author's complete address, and if possible, an electronic mail
address. Start the body of the first page 7.5 cm from the top of the
page.

The title, author names and addresses should be completely identical
to those entered to the electronical paper submission website in order
to maintain the consistency of author information among all
publications of the conference. If they are different, the publication
chairs may resolve the difference without consulting with you; so it
is in your own interest to double-check that the information is
consistent.

{\bf Abstract}: Type the abstract at the beginning of the first
column. The width of the abstract text should be smaller than the
width of the columns for the text in the body of the paper by about
0.6 cm on each side. Center the word {\bf Abstract} in a 12 point bold
font above the body of the abstract. The abstract should be a concise
summary of the general thesis and conclusions of the paper. It should
be no longer than 200 words. The abstract text should be in 10 point font.

{\bf Text}: Begin typing the main body of the text immediately after
the abstract, observing the two-column format as shown in

the present document. Do not include page numbers.

{\bf Indent}: Indent when starting a new paragraph, about 0.4 cm. Use 11 points for text and subsection headings, 12 points for section headings and 15 points for the title.

\begin{table}
\centering
\small
\begin{tabular}{cc}
\begin{tabular}{|l|l|}
\hline
{\bf Command} & {\bf Output}\\\hline
\verb|{\"a}| & {\"a} \\
\verb|{\^e}| & {\^e} \\
\verb|{\`i}| & {\`i} \\ 
\verb|{\.I}| & {\.I} \\ 
\verb|{\o}| & {\o} \\
\verb|{\'u}| & {\'u}  \\ 
\verb|{\aa}| & {\aa}  \\\hline
\end{tabular} & 
\begin{tabular}{|l|l|}
\hline
{\bf Command} & {\bf  Output}\\\hline
\verb|{\c c}| & {\c c} \\ 
\verb|{\u g}| & {\u g} \\ 
\verb|{\l}| & {\l} \\ 
\verb|{\~n}| & {\~n} \\ 
\verb|{\H o}| & {\H o} \\ 
\verb|{\v r}| & {\v r} \\ 
\verb|{\ss}| & {\ss} \\\hline
\end{tabular}
\end{tabular}
\caption{Example commands for accented characters, to be used in, {\em e.g.}, \BibTeX\ names.}\label{tab:accents}
\end{table}

\subsection{Sections}

{\bf Headings}: Type and label section and subsection headings in the
style shown on the present document.  Use numbered sections (Arabic
numerals) in order to facilitate cross references. Number subsections
with the section number and the subsection number separated by a dot,
in Arabic numerals.
Do not number subsubsections.

\begin{table*}[t!]
\centering
\begin{tabular}{lll}
  output & natbib & previous ACL style files\\
  \hline
  \citep{Gusfield:97} & \verb|\citep| & \verb|\cite| \\
  \citet{Gusfield:97} & \verb|\citet| & \verb|\newcite| \\
  \citeyearpar{Gusfield:97} & \verb|\citeyearpar| & \verb|\shortcite| \\
\end{tabular}
\caption{Citation commands supported by the style file.
  The citation style is based on the natbib package and
  supports all natbib citation commands.
  It also supports commands defined in previous ACL style files
  for compatibility.
  }
\end{table*}
\fi

%{\bf Citations}: Citations within the text appear in parentheses
%as~\cite{Gusfield:97} or, if the author's name appears in the text
%itself, as Gusfield~\shortcite{Gusfield:97}.
%Using the provided \LaTeX\ style, the former is accomplished using
%{\small\verb|\cite|} and the latter with {\small\verb|\shortcite|} or {\small\verb|\newcite|}. Collapse multiple citations as in~\cite{Gusfield:97,Aho:72}; this is accomplished with the provided style using commas within the {\small\verb|\cite|} command, {\em e.g.}, {\small\verb|\cite{Gusfield:97,Aho:72}|}. Append lowercase letters to the year in cases of ambiguities.  
 %Treat double authors as
%in~\cite{Aho:72}, but write as in~\cite{Chandra:81} when more than two
%authors are involved. Collapse multiple citations %as
%in~\cite{Gusfield:97,Aho:72}. Also refrain from using full citations
%as sentence constituents.
\if{0}
We suggest that instead of
\begin{quote}
  ``\cite{Gusfield:97} showed that ...''
\end{quote}
you use
\begin{quote}
``Gusfield \shortcite{Gusfield:97}   showed that ...''
\end{quote}

If you are using the provided \LaTeX{} and Bib\TeX{} style files, you
can use the command \verb|\citet| (cite in text)
to get ``author (year)'' citations.

If the Bib\TeX{} file contains DOI fields, the paper
title in the references section will appear as a hyperlink
to the DOI, using the hyperref \LaTeX{} package.
To disable the hyperref package, load the style file
with the \verb|nohyperref| option: \\{\small
\verb|\usepackage[nohyperref]{naaclhlt2018}|}

\textbf{Digital Object Identifiers}:  As part of our work to make ACL
materials more widely used and cited outside of our discipline, ACL
has registered as a CrossRef member, as a registrant of Digital Object
Identifiers (DOIs), the standard for registering permanent URNs for
referencing scholarly materials.  As of 2017, we are requiring all
camera-ready references to contain the appropriate DOIs (or as a
second resort, the hyperlinked ACL Anthology Identifier) to all cited
works.  Thus, please ensure that you use Bib\TeX\ records that contain
DOI or URLs for any of the ACL materials that you reference.
Appropriate records should be found for most materials in the current
ACL Anthology at \url{http://aclanthology.info/}.

As examples, we cite \cite{P16-1001} to show you how papers with a DOI

will appear in the bibliography.  We cite
\fi
%\cite{C14-1001} 
%to show how
%papers without a DOI but with an ACL Anthology Identifier will appear
%in the bibliography. 

\if{0}

As reviewing will be double-blind, the submitted version of the papers
should not include the authors' names and affiliations. Furthermore,
self-references that reveal the author's identity, {\em e.g.},
\begin{quote}
``We previously showed \cite{Gusfield:97} ...''  
\end{quote}
should be avoided. Instead, use citations such as 
\begin{quote}
``\citeauthor{Gusfield:97} \shortcite{Gusfield:97}
previously showed ... ''
\end{quote}

Any preliminary non-archival versions of submitted papers should be listed in the submission form but not in the review version of the paper. NAACL-HLT 2018 reviewers are generally aware that authors may present preliminary versions of their work in other venues, but will not be provided the list of previous presentations from the submission form.

\textbf{Please do not use anonymous citations} and do not include
 when submitting your papers. Papers that do not
conform to these requirements may be rejected without review.

\textbf{References}: Gather the full set of references together under
the heading {\bf References}; place the section before any Appendices,
unless they contain references. Arrange the references alphabetically
by first author, rather than by order of occurrence in the text.
Provide as complete a citation as possible, using a consistent format,
such as the one for {\em Computational Linguistics\/} or the one in the 
{\em Publication Manual of the American 
Psychological Association\/}~\cite{APA:83}. Use of full names for
authors rather than initials is preferred. A list of abbreviations
for common computer science journals can be found in the ACM 
{\em Computing Reviews\/}~\cite{ACM:83}.

The \LaTeX{} and Bib\TeX{} style files provided roughly fit the
American Psychological Association format, allowing regular citations, 
short citations and multiple citations as described above.

Submissions should accurately reference prior and related work, including code and data. If a piece of prior work appeared in multiple venues, the version that appeared in a refereed, archival venue should be referenced. If multiple versions of a piece of prior work exist, the one used by the authors should be referenced. Authors should not rely on automated citation indices to provide accurate references for prior and related work.

{\bf Appendices}: Appendices, if any, directly follow the text and the
references (but see above).  Letter them in sequence and provide an
informative title: {\bf Appendix A. Title of Appendix}.

\subsection{Footnotes}

{\bf Footnotes}: Put footnotes at the bottom of the page and use 9
point font. They may be numbered or referred to by asterisks or other
symbols.\footnote{This is how a footnote should appear.} Footnotes
should be separated from the text by a line.\footnote{Note the line
separating the footnotes from the text.}

\subsection{Graphics}

{\bf Illustrations}: Place figures, tables, and photographs in the
paper near where they are first discussed, rather than at the end, if
possible.  Wide illustrations may run across both columns.  Color
illustrations are discouraged, unless you have verified that  
they will be understandable when printed in black ink.

{\bf Captions}: Provide a caption for every illustration; number each one
sequentially in the form:  ``Figure 1. Caption of the Figure.'' ``Table 1.
Caption of the Table.''  Type the captions of the figures and 
tables below the body, using 11 point text.

\subsection{Accessibility}
\label{ssec:accessibility}

In an effort to accommodate people who are color-blind (as well as those printing
to paper), grayscale readability for all accepted papers will be
encouraged.  Color is not forbidden, but authors should ensure that
tables and figures do not rely solely on color to convey critical
distinctions. A simple criterion: All curves and points in your figures should be clearly distinguishable without color.

% Min: no longer used as of NAACL-HLT 2018, following ACL exec's decision to
% remove this extra workflow that was not executed much.
% BEGIN: remove
%% \section{XML conversion and supported \LaTeX\ packages}

%% Following ACL 2014 we will also we will attempt to automatically convert 
%% your \LaTeX\ source files to publish papers in machine-readable 
%% XML with semantic markup in the ACL Anthology, in addition to the 
%% traditional PDF format.  This will allow us to create, over the next 
%% few years, a growing corpus of scientific text for our own future research, 
%% and picks up on recent initiatives on converting ACL papers from earlier 
%% years to XML. 

%% We encourage you to submit a ZIP file of your \LaTeX\ sources along
%% with the camera-ready version of your paper. We will then convert them
%% to XML automatically, using the LaTeXML tool
%% (\url{http://dlmf.nist.gov/LaTeXML}). LaTeXML has \emph{bindings} for
%% a number of \LaTeX\ packages, including the NAACL-HLT 2018 stylefile. These
%% bindings allow LaTeXML to render the commands from these packages
%% correctly in XML. For best results, we encourage you to use the
%% packages that are officially supported by LaTeXML, listed at
%% \url{http://dlmf.nist.gov/LaTeXML/manual/included.bindings}
% END: remove
\section{Translation of non-English Terms}

It is also advised to supplement non-English characters and terms
with appropriate transliterations and/or translations
since not all readers understand all such characters and terms.
Inline transliteration or translation can be represented in
the order of: original-form transliteration ``translation''.

\section{Length of Submission}
\label{sec:length}

The NAACL-HLT 2018 main conference accepts submissions of long papers and
short papers.
 Long papers may consist of up to eight (8) pages of
content plus unlimited pages for references. Upon acceptance, final
versions of long papers will be given one additional page -- up to nine (9)
pages of content plus unlimited pages for references -- so that reviewers' comments
can be taken into account. Short papers may consist of up to four (4)
pages of content, plus unlimited pages for references. Upon
acceptance, short papers will be given five (5) pages in the
proceedings and unlimited pages for references. 

For both long and short papers, all illustrations and tables that are part
of the main text must be accommodated within these page limits, observing
the formatting instructions given in the present document. Supplementary
material in the form of appendices does not count towards the page limit; see appendix A for further information.

However, note that supplementary material should be supplementary
(rather than central) to the paper, and that reviewers may ignore
supplementary material when reviewing the paper (see Appendix
\ref{sec:supplemental}). Papers that do not conform to the specified
length and formatting requirements are subject to be rejected without
review.

Workshop chairs may have different rules for allowed length and
whether supplemental material is welcome. As always, the respective
call for papers is the authoritative source.

\section*{Acknowledgments}

The acknowledgments should go immediately before the references.  Do
not number the acknowledgments section. Do not include this section
when submitting your paper for review.
\fi

%\bibliography{naaclhlt2018}
%\bibliography{naaclhlt2018}
%\bibliographystyle{acl_natbib}

\if{0}
\appendix

\section{Supplemental Material}
\label{sec:supplemental}
Submissions may include resources (software and/or data) used in in the work and described in the paper. Papers that are submitted with accompanying software and/or data may receive additional credit toward the overall evaluation score, and the potential impact of the software and data will be taken into account when making the acceptance/rejection decisions. Any accompanying software and/or data should include licenses and documentation of research review as appropriate.

NAACL-HLT 2018 also encourages the submission of supplementary material to report preprocessing decisions, model parameters, and other details necessary for the replication of the experiments reported in the paper. Seemingly small preprocessing decisions can sometimes make a large difference in performance, so it is crucial to record such decisions to precisely characterize state-of-the-art methods. 

Nonetheless, supplementary material should be supplementary (rather
than central) to the paper. {\bf Submissions that misuse the supplementary 
material may be rejected without review.}
Essentially, supplementary material may include explanations or details
of proofs or derivations that do not fit into the paper, lists of
features or feature templates, sample inputs and outputs for a system,
pseudo-code or source code, and data. (Source code and data should
be separate uploads, rather than part of the paper).

The paper should not rely on the supplementary material: while the paper
may refer to and cite the supplementary material and the supplementary material will be available to the
reviewers, they will not be asked to review the
supplementary material.

Appendices ({\em i.e.} supplementary material in the form of proofs, tables,
or pseudo-code) should come after the references, as shown here. Use
\verb|\appendix| before any appendix section to switch the section
numbering over to letters.

\section{Multiple Appendices}
\dots can be gotten by using more than one section. We hope you won't
need that.
\fi

\end{document}